%% file: neurips_2024.tex
\documentclass{article}

\pdfoutput=1
    \PassOptionsToPackage{numbers, compress}{natbib}
\usepackage[preprint]{neurips_2024}

\usepackage{caption}
\usepackage[utf8]{inputenc} % allow utf-8 input
\usepackage[T1]{fontenc}    % use 8-bit T1 fonts
\usepackage{hyperref}       % hyperlinks
\usepackage{url}            % simple URL typesetting
\usepackage{booktabs}       % professional-quality tables
\usepackage{amsfonts}       % blackboard math symbols
\usepackage{nicefrac}       % compact symbols for 1/2, etc.
\usepackage{microtype}      % microtypography
\usepackage{xcolor}         % colors
\usepackage{graphicx}
\usepackage{grffile}

\title{Small Language Models are Equation Reasoners}

\author{Bumjun Kim\textsuperscript{1}
 	 \quad
	\textbf{Kunha Lee}\textsuperscript{2}
 	 \quad
	\textbf{Juyeon Kim}\textsuperscript{3}
 	 \quad
	\textbf{Sangam Lee}\textsuperscript{4}  \\
	\textsuperscript{1}Hongik University  \,  \textsuperscript{2}Sangmyung University  \,  \textsuperscript{3}Ewha Womans University   \, \textsuperscript{4}Yonsei University\\
	\textsuperscript{1}	\texttt{quasar0529@mail.hongik.ac.kr}
	\textsuperscript{2}	\texttt{kunha98@gmail.com}\\
	\textsuperscript{3}	\texttt{johnszone@ewhain.net}
	\textsuperscript{4}	\texttt{salee@yonsei.ac.kr} 
}

\begin{document}
\maketitle

\begin{abstract}
\input{000Abstract}
\end{abstract}

\section{Introduction}
\label{sec:intro}
\input{010introduction}

\section{Related Works}
\label{sec:relatedwork}
\input{020relatedworks}

\section{Experiments}
\label{sec:methods}
\input{030Methods}

\section{Conclusion}
\label{sec:result}
\input{040Result}

\cleardoublepage
\bibliography{reference}

\end{document}

%% file: 000Abstract.tex
Chain-of-Thought (CoT) reasoning has enabled Large Language Model (LLM) to achieve remarkable performance in various NLP tasks, including arithmetic problem-solving. 
However, this success does not generalize to small language model (sLM) like T5, due to their limited capacity and absence of emergent abilities associated with larger models. 
Recent works to enhance sLM through knowledge distillation have yielded some improvements but still face significant limitations, particularly high ambiguity from the variability in natural language expressions and substantial computational costs. 
In this paper, we investigate why sLM perform poorly on arithmetic reasoning tasks and hypothesize that natural language format variability introduces high ambiguity for these smaller models. 
Based on this hypothesis, we conduct experiments with equation-only format, which is a reasoning format that unifies arithmetic reasoning previously expressed in natural language formats into mathematical equations. 
Experiment results demonstrate that equation-only format effectively boosts the arithmetic reasoning abilities of sLM, especially in very small models like T5-Tiny.

%% file: 010Introduction.tex
Large Language Model(LLM)'s reasoning ability through Chain-of-Thought (CoT)~\cite{kojima2022large, wei2022chain} have demonstrated remarkable performance on various NLP downsteam tasks. 
Especially in recent times, it also has shown good results in tasks like arithmetic tasks, which involve solving mathematical problems.
However, while CoT has significantly enhanced the arithmetic performance of "Large" Language models~\cite{chowdhery2023palm,anil2023palm, achiam2023gpt, dubey2024llama}, this improvement does not generalize to small Language Model(sLM) such as T5~\cite{raffel2020exploring} due to the absence of emergent abilities, which are often linked to model scaling laws. 

While LLM offer superior performance, their tremendous computational and memory demands make it impractical for most real-world applications~\cite{Zhou2024ASO, Xu2024ASO}. 
In environments such as edge devices, mobile platforms, or real-time systems, the resources required to run these models are simply not available. 
As a result, sLM become crucial for extending the reach of advanced language technologies, offering more efficient solutions for resource-constrained settings. 
By enhancing the reasoning capabilities of sLM, we can close the gap between the high-level performance of LLM and the practical needs of real-world use cases, enabling the deployment of sophisticated reasoning models even in environments with limited computational power. 
Recent works have tried to enhance the arithmetic reasoning abilities of sLM by transferring the reasoning capabilities of LLM, through techniques such as knowledge distillation~\cite{hsieh2023distilling, zhu2024distilling, ho2023large}. 
These approaches have led to some performance improvements, but they still lack absolute performance.

To explore the potential of sLM performance on arithmetic reasoning tasks, in this paper, we hypothesize and experimentally analyze why sLM has not performed well on arithmetic reasoning tasks in existing methods. Our main hypothesis is \textbf{“Natural language format cause high ambiguity for sLM”}.
Let’s consider the mathematical expression "1+1=2." In a natural language format, this concept can be expressed in various ways. For instance, it could be framed as: "If Tom has 1 donut and Mike has 1 piece of bread, what is the total amount of food they have?" Alternatively, it could be expressed as: "If Emily has 1 MacBook and James has the same number, what is the total number of laptops they own?" This variability in natural language formats can increase ambiguity for sLM, which have relatively lower capacity compared to LLM. 

\begin{figure*}[t!]
    \centering
    \includegraphics[width=\textwidth]{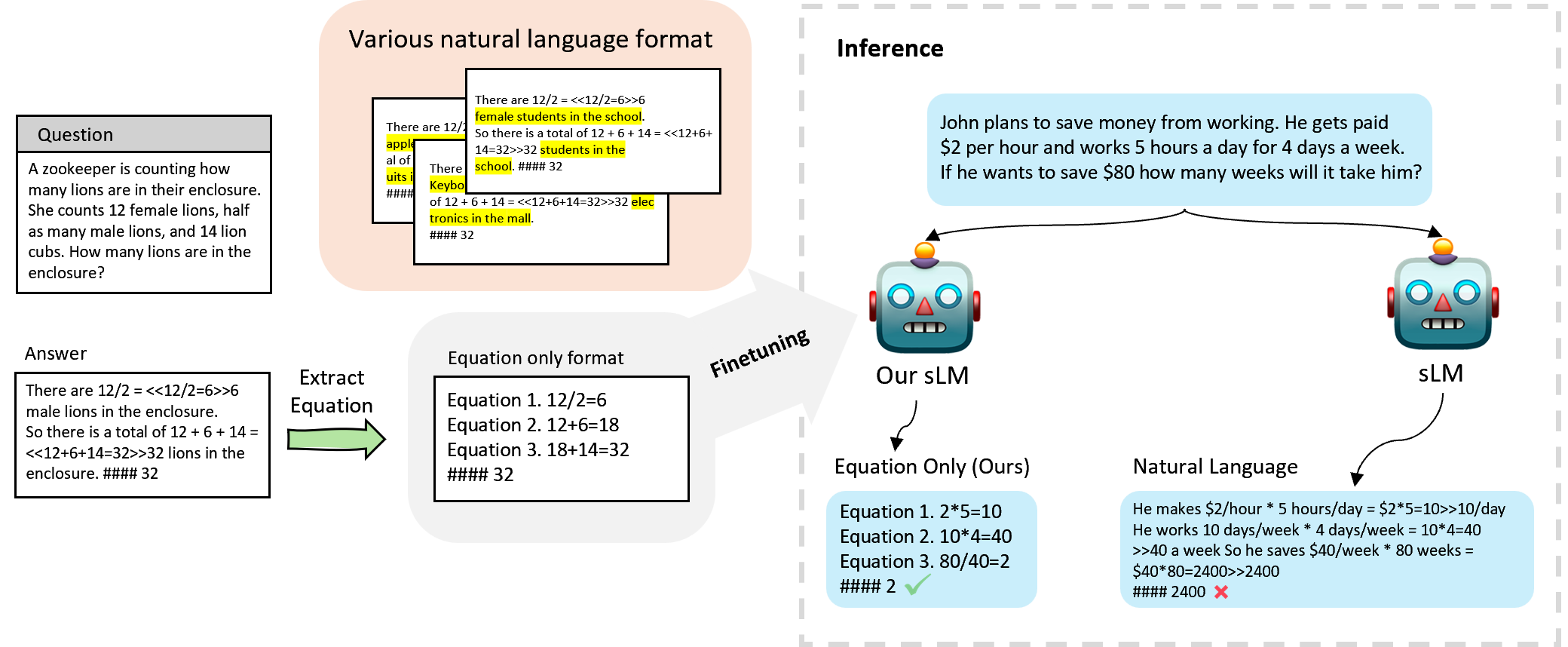}
    \caption{Overview of our experiment. We conduct experiment with two format: natural language format and equation-only format. Equation-only format corresponds to each specific mathematical problem with only a single reasoning format, eliminating variability in format and preventing sLM from experiencing ambiguity.}
    \label{fig:method}
\end{figure*}

Based on this hypothesis, we conduct experiments with \textbf{equation-only format}, which is a reasoning format that unifies arithmetic reasoning previously expressed in natural language formats into mathematical equations. 
As shown in figure \ref{fig:method}, it corresponds to each specific mathematical problem with only a single reasoning format, eliminating variability in format and preventing sLM from experiencing ambiguity. 
Our experiments have demonstrated the effectiveness of the equation-only format, showing that it is particularly effective in very small sLM like T5-Tiny with lower cost than existing methods.

%% file: 020relatedworks.tex
\subsection{Large Language Model(LLM)}
% Unsupervised pre-training on massive unlabeled-datasets, followed by task-specific fine-tuning~\cite{peters2018deep, lee2018pre} has achieved astonishing results across a wide arrange of NLP tasks.
% More recently, decoder-based generation models such as GPT/{gpt2} have demonstrated the ability that large autoregressive language model(LLM) can generally solve wide range of tasks without fine-tuning and parameter updates.

Large Language Models (LLMs), such as GPT-4~\cite{achiam2023gpt}, Llama-3~\cite{dubey2024llama} and PaLM-2~\cite{anil2023palm} have revolutionized the field of natural language processing (NLP) by significantly advancing the understanding and generation of human language. 
These LLMs demonstrate remarkable performance across diverse tasks, ranging from natural language understanding and generation to more complex reasoning tasks. 
However, due to the limitation of not being fine-tuned on task-specific datasets, LLM often performs unsatisfactorily on certain tasks in zero-shot settings~\cite{li-etal-2024-evaluating-mathematical}.
Despite these challenges, advances in prompt engineering techniques, such as in-context learning~\cite{NEURIPS2020_gpt3} and chain-of-thought reasoning~\cite{wei2022chain}, have enabled LLM to achieve state-of-the-art performance on numerous tasks without the need for additional fine-tuning.

Although LLMs have revolutionized the field of NLP, their immense computational and memory requirements make them unsuitable for many real-world applications~\cite{Zhou2024ASO, Xu2024ASO}. 
In resource-constrained environments like edge devices, mobile platforms, or real-time systems, the necessary resources to operate such large models are often unavailable. 
Consequently, sLMs play a pivotal role in bringing advanced language technologies to these settings, offering more efficient alternatives. 

\subsection{Arithmetic Reasoning}
% \textbf{LLM Reasoning} 
% Arithmetic Reasoning task is known as challenging task for language models.
% % , as even advanced models like GPT-4 have struggled to achieve strong performance in this area~\cite{achiam2023gpt}.
% Without Additional training technique~\cite{ouyang2022training} or in-context learning, LLM has difficulty with arithmetic task~\cite{patel2021nlp, cobbe2021training} which is relatively simple for humans~\cite{Li2024EvaluatingMR}.
% Recent advancements such as chain-of-thought reasoning~\cite{wei2022chain} have enabled LLM to achieve high performance on arithmetic tasks. 
% Moreover, prompt engineering methods, such as problem decomposition~\cite{zhou2022least} and self-consistency~\cite{Wang2022SelfConsistencyIC}, have allowed LLM to reach human-level performance on benchmarks like SVAMP~\cite{patel2021nlp}, GSM8K~\cite{cobbe2021training} and Math benchmark~\cite{Hendrycks2021MeasuringMP}.

Arithmetic reasoning has long been recognized as a particularly challenging task for language models. Unlike other tasks, where language models can leverage large datasets and contextual understanding, even advanced LLM have struggled with arithmetic problems without additional support. 
Recent works such as CoT~\cite{wei2022chain} reasoning, have significantly improved the performance of LLMs on arithmetic tasks. By guiding models to reason through problems step-by-step, CoT allows them to break down complex problems into more manageable parts, improving accuracy on tasks that require logical progression, including arithmetic reasoning. 
Additionally, various techniques, such as problem decomposition~\cite{zhou2022least} and self-consistency~\cite{Wang2022SelfConsistencyIC}, have further enhanced the capabilities of LLM. 
These methods have enabled LLMs to achieve near-human-level performance on established benchmarks such as SVAMP~\cite{patel2021nlp} and GSM8K~\cite{cobbe2021training}. These improvements highlight the potential of LLMs when equipped with advanced reasoning and prompting techniques. 

While LLMs have made significant strides, challenges remain. Small Language Models (sLMs), such as T5-base and GPT-2, struggle with arithmetic tasks. Chain-of-thought~\cite{wei2022chain}, which has proven crucial to improving arithmetic reasoning in LLMs, does not function effectively in sLMs due to emergent abilities at smaller scales~\cite{wei2022emergent}. 

% However, techniques like knowledge distillation (KD)~\cite{ho2023large}, which is the process of transferring knowledge from a large ‘teacher’ model to a smaller ‘student’ model enable them to mimic the reasoning patterns and performance of larger models to a degree.
% Recent advancements have sought to improve distillation techniques by using dual models~\cite{shridhar2023distilling}, or enhancing the quality of the teacher outputs~\cite{liu2024mind}, or augmenting the training dataset~\cite{zhu2024distilling}, leading to better performance of model on specific tasks. 

%% file: 030Methods.tex
\input{table/main_table}

\subsection{Experimental Setting}
\textbf{Dataset}
In order to explore how language models can effectively solve mathematical problems, we utilized the widely recognized Grade School Math 8K dataset (GSM8K)~\cite{cobbe2021training}. 
This dataset is designed to assess a model’s arithmetic reasoning and problem-solving abilities using elementary-level mathematical problems. 
As illustrated in Fig \ref{fig:method}, the task requires the model to solve math equations described in natural language and provide the correct answer to the posed question. 

\textbf{Model}
In this work, we employed the T5~\cite{raffel2020exploring} model. 
This model processes all inputs and outputs in a text format, making it well-suited for natural language tasks. For emergent abilities to activate and for performance to improve, a model needs to exceed a certain size. 
sLM is less likely to exhibit these abilities, and thus the methods commonly used in LLMs may not function as intended. 
This experiment was conducted to investigate which approaches are more suitable for small models—base, small, mini, tiny—when solving arithmetic tasks.
\subsection{Result}
As shown in Table \ref{table:main-table}, the accuracy of the T5-base model increased from 13\% to 17\%, and the T5-small model improved from 10\% to 14\%. 
A similar trend is observed in T5-mini and T5-tiny.
These results demonstrate a consistent performance improvement across all model sizes when using equations only, compared to training with the natural language format approach.
Previously, it was widely assumed that using natural language format would be more effective, regardless of model size. Because natural language is richer in information and language models are typically pre-trained on natural language datasets. 
However, the result of this experiment contradict that assumption. 
In fact, for smaller models, such as those below T5-base, it was found that using equations—symbols and numbers with consistent structure—was more effective than relying on natural language, which is inherently ambiguous. 
To examine this more closely, we compared the cross attention score of the Encoder-Decoder of the model.

\begin{figure*}[t!]
    \centering
    \includegraphics[width=\textwidth]{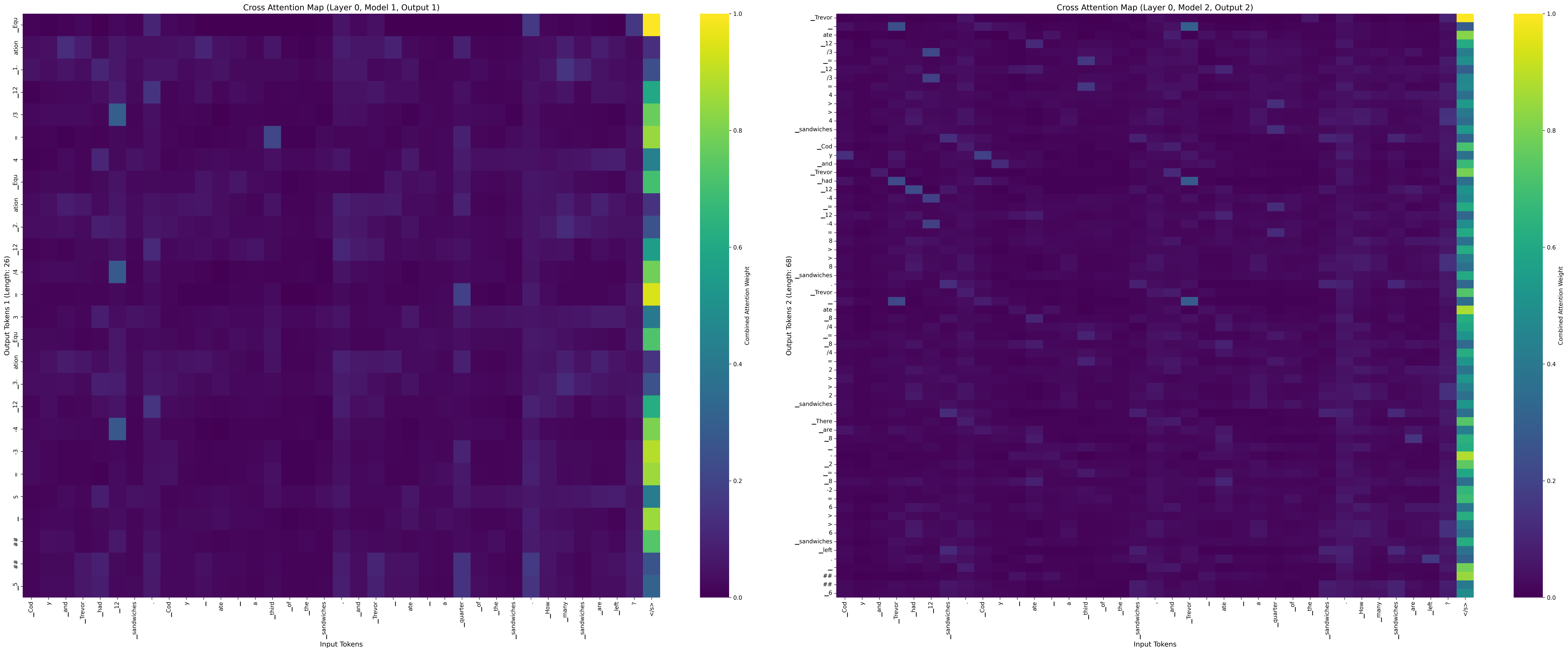}
    \caption{Cross-attention score map of T5 model}
    \label{fig:attention}
\end{figure*}

Observing the attention score map for the problems where ``Equation only'' was correct and ``Natural Language'' was incorrect in Fig \ref{fig:attention}, we found that the attention scores for paired tokens such as ``times'' and ``*'', or ``Half'' and ``/2'', were higher in equation-only format.
Furthermore, when using natural language, model generally exhibited a dispersed attention score and often assigning high scores to tokens that were unrelated to the correct answer. 
This suggests that due to the inherent ambiguity of natural language, it is necessary to consider the entire context, which may lead to a tendency to overlook truly important tokens.

%% file: table/main_table.tex
\begin{table}[!]
\centering
\renewcommand{\arraystretch}{1.5} 
\begin{tabular}{cccc}
\hline
\textbf{Model} & \textbf{ParamSize} & \textbf{Natural Language (\%)} & \textbf{Equation Only(\%)} \\ \hline\hline 
T5-Base   & 220M            & 0.13                & \textbf{0.17}                  \\
T5-Small   & 60M            & 0.10                & \textbf{0.14}                  \\
T5-Mini    & 31M            & 0.08                & \textbf{0.11}                  \\
T5-Tiny    & 16M            & 0.07                & \textbf{0.10}                  \\ \hline
\end{tabular}
\caption{Performance comparison for the Natural Language Format and Equation Only for GSM8K. We report the performance of each model per reasoning format as accuracy.}
\label{table:main-table}
\end{table}

%% file: 040Result.tex
In this paper, we investigated why small language models (sLMs) perform poorly on arithmetic reasoning tasks and proposed that the variability in natural language formats introduces significant ambiguity for these models. 
To address this, we hypothesized that by reducing the ambiguity through an equation-only format, we could improve performance. 
Our experiments demonstrated that the equation-only format consistently outperformed natural language formats, especially in smaller models like T5-Tiny, which lack the capacity to handle the inherent ambiguity of natural language reasoning effectively. In equation-only format, it was observed in attention score map that various names of variable and operation symbols were better mapped than natural language format.

Finding of this work suggests that simplifying reasoning tasks into more structured formats like equations can significantly enhance the arithmetic capabilities of sLMs without increasing computational costs. 
This is especially beneficial in resource-constrained environments where large models like LLMs are impractical. 
By adopting such methods, sLMs can be better optimized for real-world applications, making advanced reasoning more accessible and efficient. 
Future work could explore the application of this approach to other reasoning tasks, potentially expanding the utility of sLMs in various domains.